\documentclass{article}

\usepackage{microtype}
\usepackage{graphicx}
\usepackage{booktabs} 
\usepackage{amsmath,amsthm,amsfonts,amssymb,amscd}
\usepackage{lastpage}
\usepackage{enumerate}
\usepackage{fancyhdr}
\usepackage{mathrsfs}
\usepackage{xcolor}
\usepackage{hyperref}
\usepackage{mathtools}

\usepackage{amssymb}
\usepackage{subcaption}
\usepackage{array}

\usepackage{stackengine} 

\usepackage{amsmath}
\usepackage{bbm}
\DeclareMathOperator*{\argmax}{argmax}

\DeclareMathOperator*{\real}{\rm I\!R}
\usepackage{listings}
\usepackage{dsfont}

\usepackage{hyperref}

\usepackage{standalone}

\usepackage[accepted]{icml2020}

\icmltitlerunning{GACEM: Generalized Auto-regressive Cross Entropy Method for Multi-Modal constraint Satisfaction Problem}

\begin{document}

\twocolumn[
\icmltitle{GACEM: Generalized Autoregressive Cross Entropy Method \\ for Multi-Modal Black Box Constraint Satisfaction}



\icmlsetsymbol{equal}{*}

\begin{icmlauthorlist}
\icmlauthor{Kourosh Hakhamaneshi}{from}
\icmlauthor{Keertana Settaluri}{from}
\icmlauthor{Pieter Abbeel}{from}
\icmlauthor{Vladimir Stojanovic}{from}

\end{icmlauthorlist}

\icmlaffiliation{from}{University of California Berkeley}

\icmlcorrespondingauthor{kourosh}{kourosh\_hakhamaneshi@berkeley.edu}

\icmlkeywords{Black-box Optimization, CEM, Neural Networks, Machine Learning, ICML}

\vskip 0.3in
]




\begin{abstract}

In this work we present a new method of black-box optimization and constraint satisfaction. Existing algorithms that have attempted to solve this problem are unable to consider multiple modes, and are not able to adapt to changes in environment dynamics. To address these issues, we developed a modified Cross-Entropy Method (CEM) that uses a masked auto-regressive neural network for modeling uniform distributions over the solution space. We train the model using maximum entropy policy gradient methods from Reinforcement Learning. Our algorithm is able to express complicated solution spaces, thus allowing it to track a variety of different solution regions. We empirically compare our algorithm with variations of CEM, including one with a Gaussian prior with fixed variance, and demonstrate better performance in terms of: number of diverse solutions, better mode discovery in multi-modal problems, and better sample efficiency in certain cases.
\end{abstract}

\section{Introduction}
Black-box optimization has been regarded as an alternative to gradient-based optimizations in many complex Reinforcement Learning tasks, primarily because of better scaling with computational resources \cite{salimans_evolution_2017}. Within this context there are problems where access to even estimates of gradients is expensive in terms of either run-time or feasibility. Examples include hyper-parameter optimization \cite{bergstra_random_nodate}, circuit design \cite{hakhamaneshi_bagnet_2019, settaluri_autockt_2020, zhang_circuit-gnn_2019}], and reliability estimation \cite{de_boer_tutorial_2005}.

This work addresses the problem of finding solutions to a constraint satisfaction problem where evaluation of constraint metrics is expensive. Formally, let $f_i(\mathbf{x}): \chi \to \real$, $i = 1, ..., m$ be $m$ functions evaluated over some design space $\chi \subset \real^d$. The goal is to find all $\mathbf{x}$ such that $f_i(\mathbf{x}) \le 0$. This formalization also encompasses the classical optimization definition where the goal is to $\min_\mathbf{x} g(\mathbf{x})$ subject to $f_i(\mathbf{x}) \le 0$ for $i=1,...,m$. We only need to set $f_{m+1}(\mathbf{x}) = g(\mathbf{x}) - \gamma$, where $\gamma$ is a threshold that incrementally gets smaller until no solution is found.

In this work, a solution is treated as a realization of a random vector $\mathbf{X}$ defined on $\chi$ with some probability mass $p(\mathbf{x})$. The solution we would like to find puts uniform probability on the regions of $\chi$ that satisfy all the constraints (i.e. $p(\mathbf{x}) \sim \mathcal{U}$ if $\mathbf{x} \in \{\mathbf{x}| f_i(\mathbf{x}) \le 0 ,\ \forall i\}$). Solution instances can then be accessed by simply sampling $p(\mathbf{x})$. The advantage of learning such a distribution is that, in many cases, the learned solution regions from this algorithm can be more easily adapted to changes in environment dynamics through fine-tuning, as the algorithm is not greedy. To practically illustrate, in circuit design problems, there exist cheap but inaccurate simulation environments that can be used to reflect the general structure of the solution space. Expensive and accurate simulations are then used to refine the boundaries of the learned distributions with potentially less number of evaluations compared to learning everything via the expensive simulator. In robotics, simulation environments are accessible and fast to run, but real-time experiments on physical robots are time consuming. In such cases, a greedy optimizer will fail to adapt to the deployment environment due to changes in dynamics. 

For simplicity, we focus on learning the distribution when there is only one active constraint $f(\mathbf{x})$. To generalize to scenarios with multiple constraints, one can create a new constraint $f(x) = \max_if_i(x) \le 0$ and find the solution distribution for constraint $f(x) \le 0$. 

In this work, we present a generalized cross-entropy method, with the specific goal of solving multi-modal optimization problems and diversity in solutions. We demonstrate that our algorithm has the capability of exploring several modes in the solution space simultaneously, has improved scalability to higher dimensions, and better solution diversity. The main features of our approach are:
\begin{itemize}
    \item The use of expressive neural networks to model uniform unstructured distributions.
    \item An adaptive sampling scheme for training, inspired by CEM and policy gradient reinforcement learning.
\end{itemize}

For the rest of this article, in section \ref{sec: background}, background information will be provided, in section \ref{sec: tech-info} the algorithm will be proposed, and in Section \ref{sec: experiments} the performance of the proposal will be empirically investigated compared to traditional CEM.

\section{Background}
\label{sec: background}
This section will give a background and overview of related literature in this area. In section \ref{sec: EDAs}, the cross-entropy method and other similar approaches will be discussed. Section \ref{sec:Explicit Generative models} covers some of the state of the art generative models.
\subsection{Estimation of Distribution Algorithms (EDAs)}
\label{sec: EDAs}
Evolutionary algorithms optimize a function by maintaining a population of solution samples that are generated in an adaptive manner. Starting from a random initial population, these set of algorithms generate new individuals in the vicinity of previous elite samples. Some of these algorithms are ad-hoc, heuristic-based optimization methods like genetic algorithms, while some others are based on estimating a distribution over elite samples (e.g. Estimation Distribution Algorithms (EDAs)). Cross-Entropy Method (CEM), and Covariance Matrix Adaptation Evoluionary strategy (CMA-ES) are among the most well-known algorithms in this category. In the simplest form of CEM, the underlying distribution is assumed to be a multivariate gaussian, and mean and covariance matrix of its probability density function (PDF) are adjusted to increase the likelihood of sampling more individuals similar to the elite ones in the next iterations. Specifically, the update rule is:

\begin{equation}
    \hat{\mu} = \sum_{\mathbf{x}_i \in \text{Elites}} \lambda_i \mathbf{x}_i
\end{equation}
\begin{equation}
    \hat{\Sigma} = \sum_{\mathbf{x}_i \in \text{Elites}} \lambda_i (\mathbf{x}_i - \hat{\mu})(\mathbf{x}_i - \hat{\mu})^T + \epsilon I_{d\times d}
\end{equation}

$\lambda_i$ are the weights of each individual and can be equal or different depending on their performance \cite{hansen_cma_2016}.
$\epsilon$ is an added noise factor with zero mean whose variance decays from $\sigma_{init}$ to $\sigma_{end}$ to encourage exploration. At each iteration $n_s$ samples are obtained from the distribution and evaluated through the constraint function. The top $q$ percentile of them are then chosen as elites. In this work we will empirically show that not adapting the co-variance will provide more diversity in solution and exploration. This method was originally developed as a solution to rare event estimation \cite{rubinstein_optimization_1997}, and was extended to support optimization problems \cite{rubinstein_cross-entropy_2002}. There is an unknown distribution that will only generate samples of that rare event (hence giving the most sample efficient estimation). CEM will minimize the KL divergence between a Gaussian prior and that optimal distribution in an adaptive sampling scheme.

CEM has the following drawbacks:

\begin{enumerate}
    \item The original implementation is very greedy, as it makes the co-variance matrix converge to zero (i.e. $E[\hat{\Sigma}] \to 0$). This property is not limiting when the goal is to find the optimal solution, but modifications are needed in order to find a high entropy solution distribution.
    
    \item The parametrized distribution family does not have the capacity to model arbitrary multi-modal (and uniform) distributions. 
    
    \item This approach will fail when scaling to high dimensional problems. This is discussed in \cite{geyer_cross_2019} in the context of reliability estimation problems. 
\end{enumerate}

In this work we use expressive generative neural networks (Section \ref{sec:Explicit Generative models}) to address the incapability in modeling multi-modal distributions. To train the model, we apply a similar approach to policy gradients in Reinforcement Learning for updating parameters.

\subsection{Explicit Generative Models}
\label{sec:Explicit Generative models}
Explicit generative models leverage Neural Networks (NN) to model the joint probability distribution of the input variables, and are optimized based on the principle of maximizing likelihood over the data. They are generally categorized into 3 classes: latent variable models, flow models, and auto-regressive models.

Latent variable models construct a graphical model over hidden variables and use probabilistic inference to maximize the variational lower bound of the joint distribution \cite{kingma_auto-encoding_2014}. In this work we are interested in maximizing the likelihood. However, these models only optimize an approximate likelihood and thus latent variable models cannot be easily used.

Flow models represent the data as a reversible non-linear mapping from a random variable with known distribution to data. To sample such a model, the random variable is generated and fed through the forward pass of the NN. To compute the probability, the data is fed through the backward pass and queried (i.e. \cite{dinh_density_2017, kingma_glow_2018}). 

In auto-regressive models, the joint distribution is factorized into partial conditional distributions using Bayes rule. Each term is modeled as the output of a single NN where the dependency of output to the inputs is carefully designed to account for conditional distribution dependencies (i.e. MADE \cite{germain_made_2015}, PixelCNN++ \cite{salimans_pixelcnn_2017}). 

In this work, we use auto-regressive models because they generally have better likelihood estimates than flow models. Our adaptive sampling scheme, however, can be applied to any other model that optimizes likelihood.

\begin{figure}
    \centering
    \includegraphics[width=\linewidth]{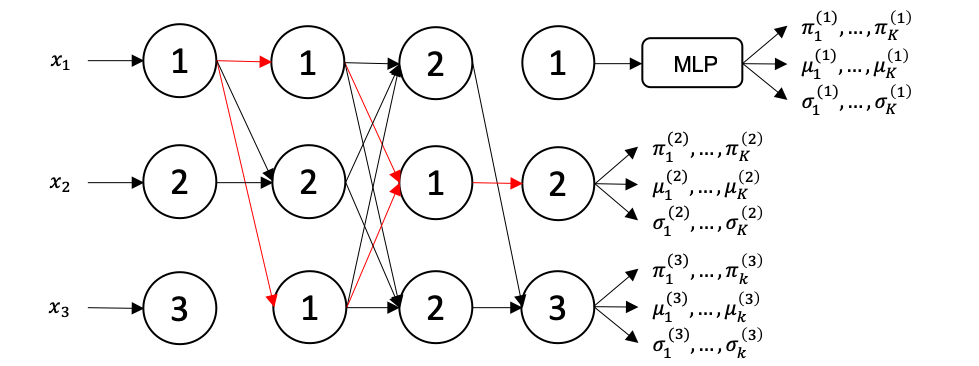}
    \caption{NN architecture of the masked auto-encoder, red arrows: dependencies of unit 2, black arrows: dependencies of unit 3}
    \label{fig:MADE_arch}
\end{figure}

\section{Generalized CEM for Constraint Satisfaction}
\label{sec: tech-info}
In the following section, we discuss the details of our algorithm. Section \ref{sec: made} introduces the masked auto-encoder architecture we use as the auto-regressive model. Our training algorithm is presented in section \ref{sec: RL}. Psuedocode is presented for clarity.



\subsection{Masked Auto-encoder Architecture}
\label{sec: made}

In this work, we use a Masked Auto-encoder for Distribution Estimation (MADE) model with some modifications from PixelCNN++. As mentioned before, by doing so, we address the difficulty in exploring multi-modal problems. 

The joint distribution is written as follows:
{
\footnotesize
\begin{equation}
    p_{\theta}(\mathbf{x}=(x_i)_{i=1}^{n}) = p_{\theta}(x_1)p_{\theta}(x_2|x_1)...p_{\theta}(x_n|x_{n-1}...x_1)
\end{equation}
}
In MADE each $p_{\theta}(x_i|.)$ is specified as a softmax with $p$ outputs that sum to one ($p$ is the number of possible values each variable $x_i$ can take). 
In this work each $p_{\theta}(x_i|.)$ is modeled as a mixture of Gaussian functions (i.e $p_{\theta}(x_i|.) = \sum_{k=1}^{K}\pi_k \mathcal{N}(x_i; \mu_k, \sigma_k)$, with $K$ number of mixtures where $\pi_k(.), \mu_k(.), \sigma_k(.)$ depend on conditional variables for $x_i$). For discrete environments this distribution is on a continuous latent variable $\nu$ which is discretized to $x_i$'s domain. Hence, this approach, supports both continuous and discrete domain optimizations. 
Figure \ref{fig:MADE_arch} shows the architecture of a simplified model with 3 design variables. Note that the output of unit one only depends on the bias terms from the second to last layer, while unit two depends only on $x_1$ (red arrows), and unit three on both $x_1$ and $x_2$ (black arrows). For more flexible optimization of $p_\theta(x_1)$ we add an extra Multi Layer Perceptron (MLP) module to the output of unit one, otherwise the only parameters for adjusting that probability are the bias terms in second to last layer.  

For each discrete design variable we associate a continuous random variable $\nu$ with probability distribution $p_\theta(\nu) = \frac{\sum_{k=1}^K \pi_k \mathcal{N}(\nu; \mu_k, \sigma_k)}{\sum_{k=1}^K \pi_k}$. Then, a uniform mapping is set from possible discrete values to the range of $[-1,1]$ (i.e. $x \in \chi \to \tilde{x} \in [-1, 1]$ with bin intervals of $\delta$). Then the probability of $x$ is the normalized version of Eq. \ref{eq: discrete prob}.  We normalize to make sure that the sum is still one. For continuous design variables we use the distribution of $\nu$ directly.

In this architecture, fixing the variance is still allowed to provide more exploration and easier optimization. In that case, each node in Figure \ref{fig:MADE_arch} will output the mean and summation coefficients of mixtures of Gaussians to adjust the PDF. This way of modeling, will allow the expression of more complicated and multi-modal distributions, as opposed to CEM.

\begin{equation}
    \footnotesize
    \label{eq: discrete prob}
    p(x) \propto \sum_{k=1}^K \pi_k \mathcal{N}(\tilde{x}+\delta/2; \mu_k, \sigma_k) - \pi_k \mathcal{N}(\tilde{x}-\delta/2; \mu_k, \sigma_k)
\end{equation}

\subsection{Training with REINFORCE}
\label{sec: RL}
In CEM, the choice of variance is arbitrary, and even if the variance is adapted (i.e. CMA-ES), it can become too greedy and results in limited exploration of space. We overcome this issue by taking entropy into account during the optimization process, and simultaneously exploring various regions of the space.

EDA algorithms represent the population of solutions with a distribution over problem variables $p_\theta(\mathbf{x})$ (itself parameterized by $\theta$) and their objective is to maximize the likelihood of elite individuals with respect to $\theta$ using stochastic gradient ascent. The estimate of the gradient can be derived in a very similar manner to REINFORCE \cite{williams_simple_1992}. Let $p_\theta(x)$ present the uniform distribution in the region that $f(x) \le 0$ where $f(.)$ is the constraint metric of the optimization problem. The objective is to find a maximum entropy distribution $p_\theta(x)$ that also maximizes the reward $w(x)$. Eq. \ref{eq: mature} formalizes the objective, with $\beta$ being a tuning factor that balances randomness with accuracy of the distribution. Another interpretation for Eq. \ref{eq: mature} is that for a given $\beta$ there is an $H_0$ that sets a minimum entropy requirement for the optimization problem in the form of Eq. \ref{eq: constraint_opt}.

\begin{equation}
    \label{eq: mature}
    W(\theta) = E_{x\sim p_\theta} [w(x)] + \beta H(p_\theta)
\end{equation}

\begin{align}
    \label{eq: constraint_opt}
\begin{split}
    \argmax_\theta W(\theta) = &\argmax_\theta  E_{x\sim p_\theta} [w(x)] \\
    & \text{s.t.} \ H(p_\theta(x)) \ge H_0
\end{split}
\end{align}

Similar to REINFORCE we can compute an unbiased estimate of the gradient in the following manner:

{
\footnotesize
\begin{align*}
    \nabla_{\theta}W(\theta) = E_{x\sim p_\theta} [\underbrace{(w(x) - \beta (1+\log p_\theta(x)))}_{\tilde{w}(x;\theta)}\nabla_\theta \log p_{\theta}(x)]
\end{align*}
\begin{align}
    \label{eq: on-policy}
    \hat{\nabla_{\theta}W(\theta)} = \frac{1}{N} \sum_{i=1}^N \tilde{w}(x_i;\theta)\nabla_\theta \log p_{\theta}(x_i)
\end{align}
}
Where $x_i \sim p_\theta(x)$ or with important sampling we can use any sampling distribution $x_i \sim p_{\theta'}(x)$:

{
\footnotesize
\begin{equation}
    \label{eq: important_sampling}
    \hat{\nabla_{\theta}W(\theta)} = \frac{1}{N} \sum_{i=1}^N \tilde{w}(x_i;\theta)\nabla_\theta \log p_{\theta}(x_i) \frac{p_{\theta}(x_i)}{p_{\theta'}(x_i)}
\end{equation}
}

\subsection{The choice of $w(x)$}
Originally the CEM method uses an indicator function as $w(x)$ in Eq. \ref{eq: mature} (
$I(x) = \begin{cases}
1 & f(x) \le \gamma^{(l)}\\
0 & o.w.
\end{cases}$ where $\gamma ^ {(l)}$ is the threshold that gets smaller after each iteration until it hits zero. Another alternative is to use a smooth continuous function that also gets updated as it progresses, but the values are now informative about the relative performance of individuals. 

In this work, we use the following function: in the off-policy case (that utilizes a replay buffer) let's  call the performance with rank $m$ (to be determined by user), $\bar{f}^{(l)}$. For the on-policy case, rank $m$ is chosen among individuals in the current batch of samples. $w(x)$ can be expressed in the following manner:
{
\footnotesize
\begin{equation}
    \label{eq: weight}
    w(x; \bar{f}^{(l)}) = \begin{cases}
    1 & f(x) \le 0 \\
    exp\{-\frac{d(f(x), 0)}{d(f(x) , \bar{f}^{(l)})}\} & 0 \le f(x) \le \bar{f}^{(l)} \\
    -exp\{-\frac{1}{d(f(x) , \bar{f}^{(l)})}\} & f(x) > \bar{f}^{(l)}
    \end{cases}
\end{equation}
}
where $d(x, y)$ is a distance measure (i.e. euclidean distance) of how far $x$ is from the target value $y$. This way of expressing $w(x)$ will adjust itself if rank $m$ performance is improved. It is always 1 for those samples satisfying the constraint, between 0 and 1 for those better than rank $m$, and between -1 and 0 for those with worse than the rank $m$ performance.


\subsection{Importance Sampling}
On-policy sampling can be used to obtain new samples. The algorithm however, will become greedy; as soon as it finds a region with many positive $w(.)$ values, it will only generate samples from there. To address this issue, we draw samples from the replay buffer to compute the gradient, and importance sampling can be used to compensate for mismatch in distributions. For $p_{\theta'}$ in Eq. \ref{eq: important_sampling} another MADE model can be used to fit a distribution to samples in the buffer at each iteration. Therefore, at each iteration $l$, $p_{\theta'}$ gets updated with samples in the replay buffer.

\subsection{Pseudo-code}
Algorithm \ref{alg:GACEM} summarizes our off-policy methodology. For on-policy, Eq. \ref{eq: on-policy} can be used for gradient estimates to avoid using an extra model $p_{\theta'}$. One can use the difference between $p_{\theta^ {(t)}}(x)$ and $p_{\theta^{(t-t_0)}}(x)$ for some $t_0$ being less than some threshold as the stopping criteria. Distance measures can be estimates of KL divergence or Wasserstein distance. Our implementation uses maximum number of iterations instead, for easier implementation.
\begin{equation*}
    D(p_{\theta^ {(t)}}(x), p_{\theta^{(t-t_0)}}(x)) < \epsilon_{th}
\end{equation*}

\begin{algorithm}[tb]
   \caption{GACEM with off-policy training}
   \label{alg:GACEM}
\begin{algorithmic}
    \STATE {\bfseries Input:} constraint $f(x) \le 0$, Design space $\chi$, number of epochs updates per iterations $n_{e}$, $n_{init}$ samples for initialization, $n_s$ samples per iteration
    \STATE {\bfseries Initialize:} parameter $\theta$ for uniform solution dist., and $\theta'$ for uniform replay buffer dist.; sample and evaluate $n_{init}$ samples from $p_{\theta}$ and initialize replay buffer $\mathcal{B}$
    \WHILE{stop condition not met}
    \FOR{epoch=0; epoch $ < n_e$}
    \STATE Update parameters of $p_{\theta'}(x)$ with samples from $\mathcal{B}$
    \STATE Update parameters of $p_{\theta}(x)$ with samples from $\mathcal{B}$ according to Eq. \ref{eq: important_sampling}, with computed weights $w(x, \bar{f}^{(l)})$
    \ENDFOR
    \STATE sample and evaluate unseen individuals $\mathbf{x}^{(1)}, ...., \mathbf{x}^{(n_s)}$ from $p_{\theta}(x)$ and add them to $\mathcal{B}$
    \ENDWHILE
\end{algorithmic}
\end{algorithm}

\section{Experiments}
\label{sec: experiments}
In this section, different experiments are conducted to evaluate particular aspects of algorithm \ref{alg:GACEM} and its on-policy counterpart, relative to CEM baselines.\footnote{code available at \\ \href{https://rebrand.ly/gacem_icml2020}{\texttt{https://rebrand.ly/gacem\_icml2020}}}.
In this section we answer the following questions:
\begin{enumerate}
    \item How do the algorithms perform in multi-modal scenarios and how well do they learn the shape of an arbitrary solution space?
    \item How do they compare in terms of exploration?
    \item How do they perform with increased number of dimensions?
\end{enumerate}

\newcolumntype{F}{ >{\centering\arraybackslash} m{0.2\linewidth} }
\newcolumntype{C}{ >{\centering\arraybackslash} m{0.05\linewidth} }

\begin{figure*}
    \centering
    \resizebox{\linewidth}{!}{
    \begin{tabular}{CFFFF}
         \textbf{Name} &  \textbf{Ackley} & \textbf{Styblinski} & \textbf{Levy} & \textbf{Synt} \\
         \textbf{2D} & \includegraphics[width=\linewidth]{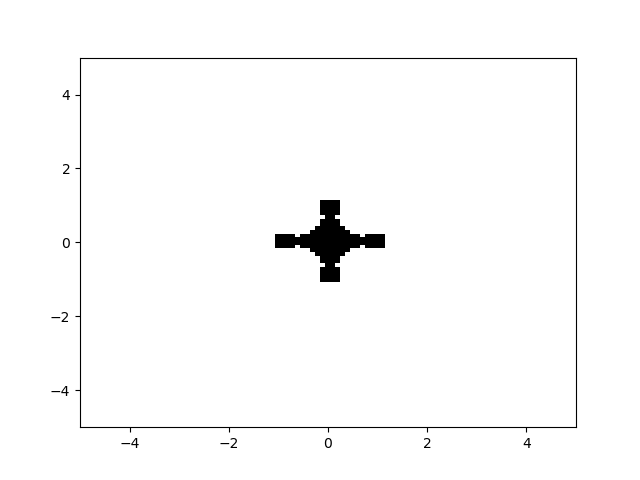} & \includegraphics[width=\linewidth]{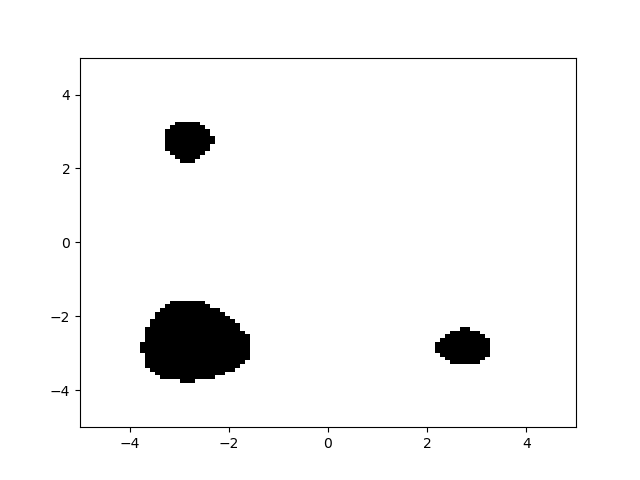} & \includegraphics[width=\linewidth]{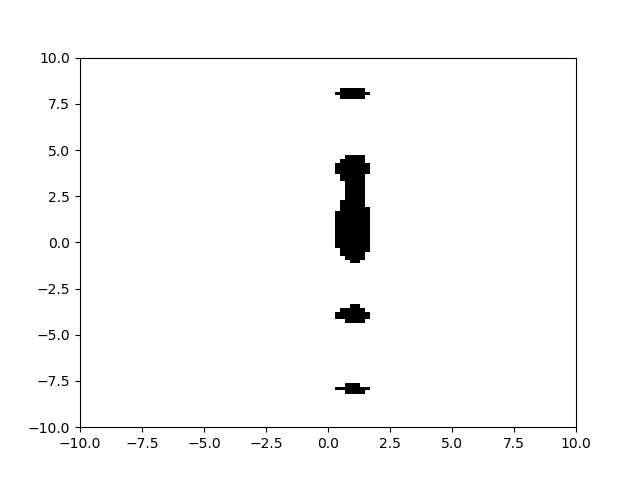} & 
         \includegraphics[width=\linewidth]{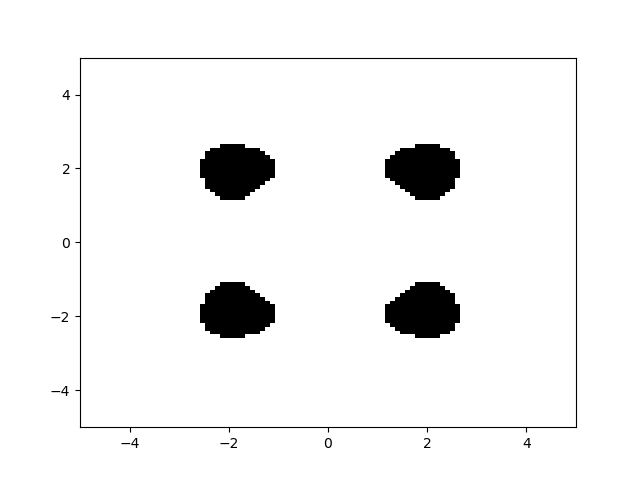}\\
         \textbf{Goal} & $f(x) < 3.5 $ & $f(x) < 20$ & $f(x) < 0.4$ & $f(x) < 2$\\
    \end{tabular}}
    \caption{Test Functions: \textbf{Ackley} ( $-20\exp(-0.2\sqrt{\frac{1}{d}\sum_{i=1}^dx_i^2)} - \exp({\frac{1}{d}\sum_{i=1}^d{cos(2\pi x_i)}}) + 20 + \exp(1)$), \textbf{Styblinski} ($\frac{1}{d}\sum_{i=1}^d(x_i^4 - 16x_i^2 + 5x_i) + 50$), \textbf{Levy} ($sin^2(\pi \omega_1) + \sum_{i=1}^{d-1}(w_i - 1)^2[1+10sin^2(\pi\omega_i + 1)] + (w_d-1)^2[1+sin^2(2\pi\omega_d)]$, where $\omega_i = 1 + \frac{x_i - 1}{4}$)} \textbf{Synt} ($\frac{1}{d}\sum_{i=1}^d\frac{1}{4}x_i^4-2x_i^2 + 5$)
    \label{fig: objectives}
\end{figure*}

\begin{figure*}
\begin{subfigure}{0.49\linewidth}
    \includegraphics[width=\linewidth]{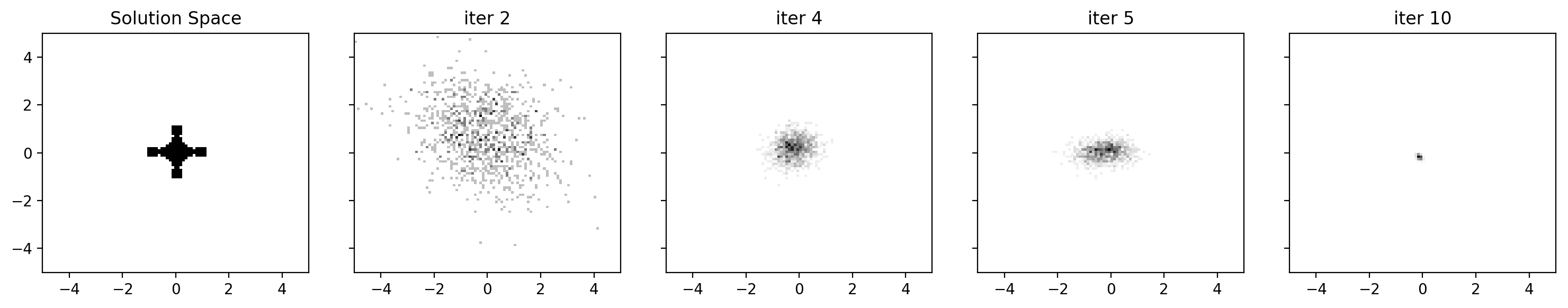}
    \par
    \includegraphics[width=\linewidth]{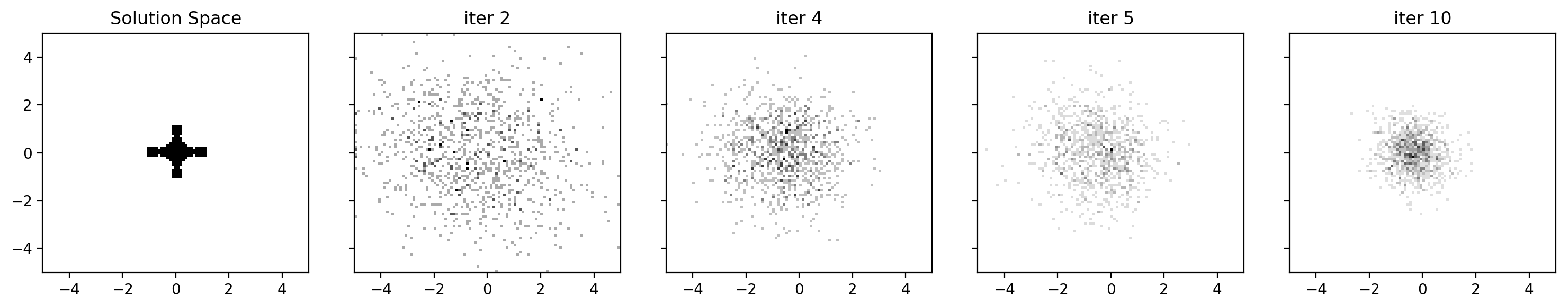}
    \caption{}
    \label{fig: cem_cempp_greedy}
\end{subfigure}
\begin{subfigure}{0.49\linewidth}
    \includegraphics[width=\linewidth]{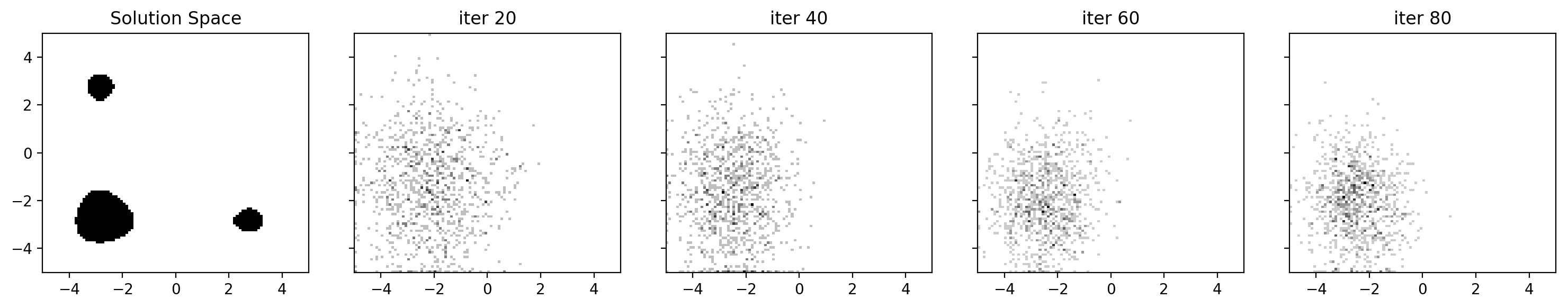}
    \par
    \includegraphics[width=\linewidth]{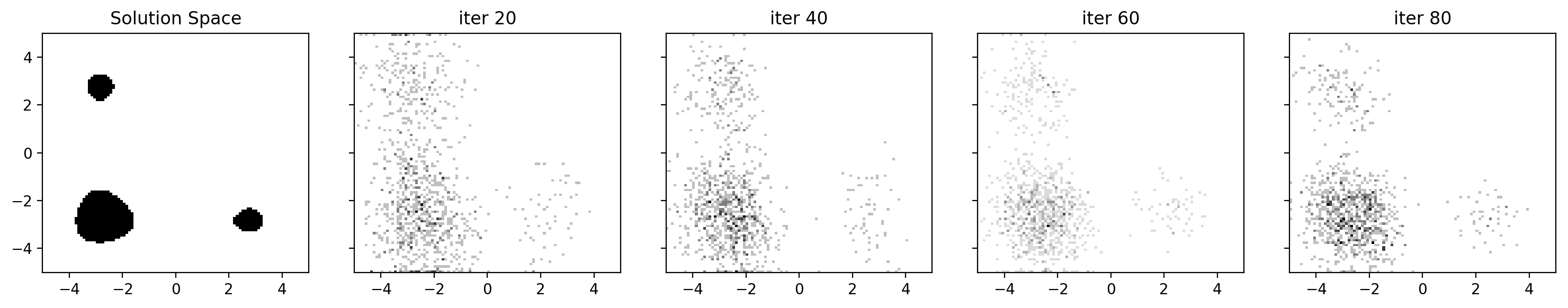}
    \caption{}
    \label{fig: cempp_multi_modal}
\end{subfigure}
\caption{Exemplary distribution development in the process of some variants of CEM. a) Comparison on Ackley function (solution plotted on left) between CEM with adaptive variance (top) and CEM++SG (bottom) b) Comparison on Styblinski function between CEM++SG (top) and CEM++KDE (bottom) }
\label{fig: dist-dev}
\end{figure*}

\subsection{Experiment Setup}
\label{sec: exp_setup}
A list of reference optimization problems are considered in Figure \ref{fig: objectives}.
The effect of higher dimensions in the design space can be easily assessed using these functions, as they are parameterized by their dimensionality. We test these algorithms on four different objectives: Ackley in 2D has a uni-modal solution space with an arbitrary shape, while Styblinksi and Levy have a multi-modal solution space. Synt is a synthetic function, useful for evaluating mode discovery. This function has $2^d$ minimums of the form $x^* = (\pm2, ..., \pm 2)$, with a value of 1. A constraint of $f(x) < 2$ will result in $2^d$ clouds of answers that needs to be discovered.
The hyper-parameters are kept constant across all experiments, showing the robustness of the algorithm. The details are provided in the supplementary materials. 

\subsection{Baselines}
\label{sec: baseline}
In addition to comparing against CEM, we also provide other CEM-based baseline algorithms that address the greedy nature and lack of expressiveness of CEM (CEM++SG, CEM++KDE).

Figure \ref{fig: cem_cempp_greedy} (top) shows an exemplary development of the distribution for finding a uni-modal 2D solution space, using CEM with adaptive variance. Each plot shows the distribution at a given iteration step shown on the title.  Without any constraints on the variance, the algorithm will make it converge to zero (even if the goal is to satisfy a constraint and not to fully optimize it). This variance minimization is due to the on-policy nature of CEM. To address this, we can fix (or schedule) the variance and not allow it to adapt (this will be referred to as CEM\_Fixed\_Variance). Alternatively, In CEM++, samples obtained at each iteration are added to a buffer, from which the mean and variance of the distribution are estimated based on the top performing samples in that buffer. Figure \ref{fig: cem_cempp_greedy} (bottom) illustrates CEM++'s distribution development for the same problem.

To cope with the second issue, instead of assuming a Single Gaussian (SG) PDF for the solution space, Kernel Density Estimation (KDE) \cite{kde_scott} is used to estimate the PDF of the top performing samples in the buffer. Figure \ref{fig: cempp_multi_modal} shows an exemplary distribution development for a multi-modal solution space, using CEM++SG (top) vs. CEM++KDE (bottom). 


CEM++SG is less greedy and provides more exploration. However in multi-modal problems it is more susceptible to getting stuck between two modes compared to the on-policy CEM. CEM++KDE provides more capacity in finding multi-modal solution spaces. As we will see in the experiments, CEM++ will not scale well to high dimensional problems. The accuracy of solutions also decreases as we move away from CEM to CEM++SG and then to CEM++KDE. This is due to the distribution estimates being smoother and containing more non-satisfactory regions with non-zero probability mass. For CEM\_Fixed\_Variance the accuracy highly depends on the choice of the variance. Smaller variance will result in greedier solution space and therefore, higher accuracy. 

\begin{figure}
\begin{subfigure}{\linewidth}
    \includegraphics[width=\linewidth]{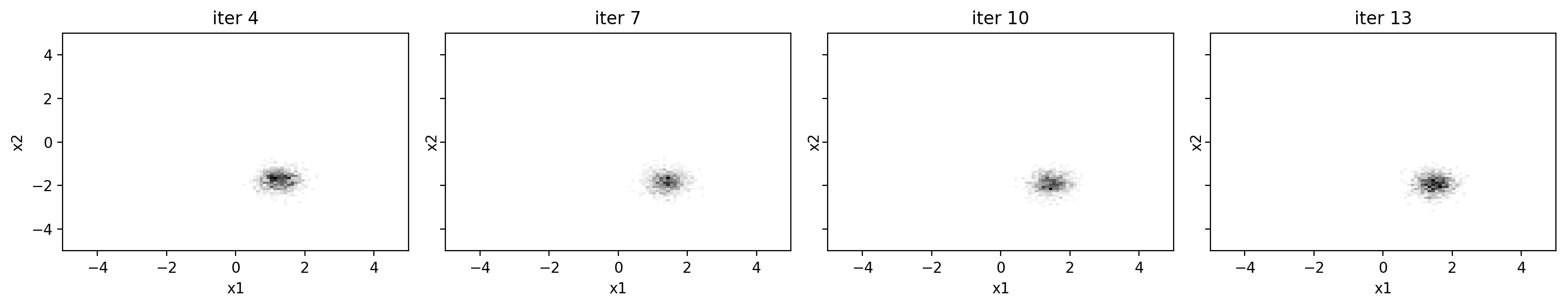}
    \caption{} \par
    \includegraphics[width=\linewidth]{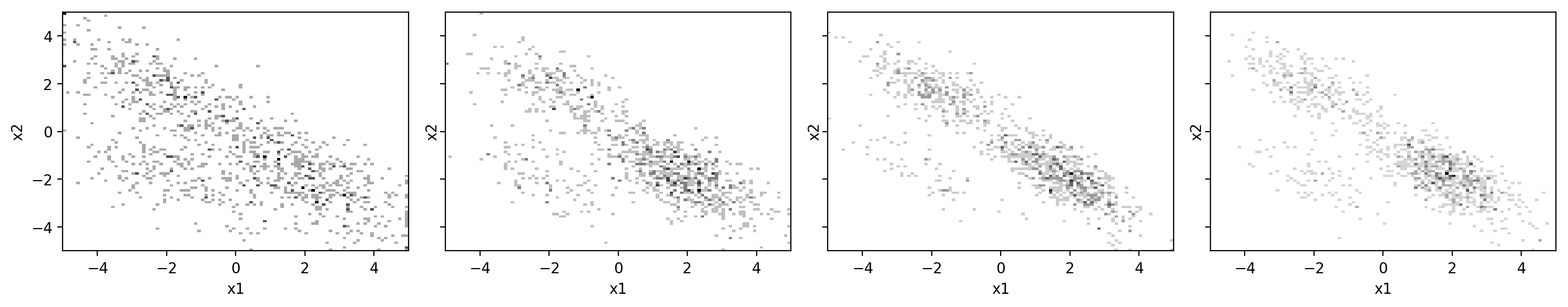}
    \caption{} \par
    \includegraphics[width=\linewidth]{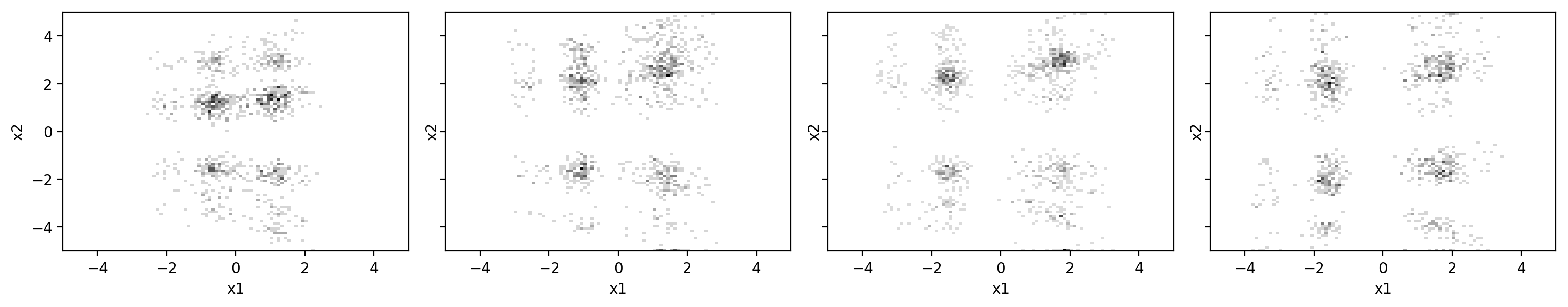}
    \caption{} \par
    \includegraphics[width=\linewidth]{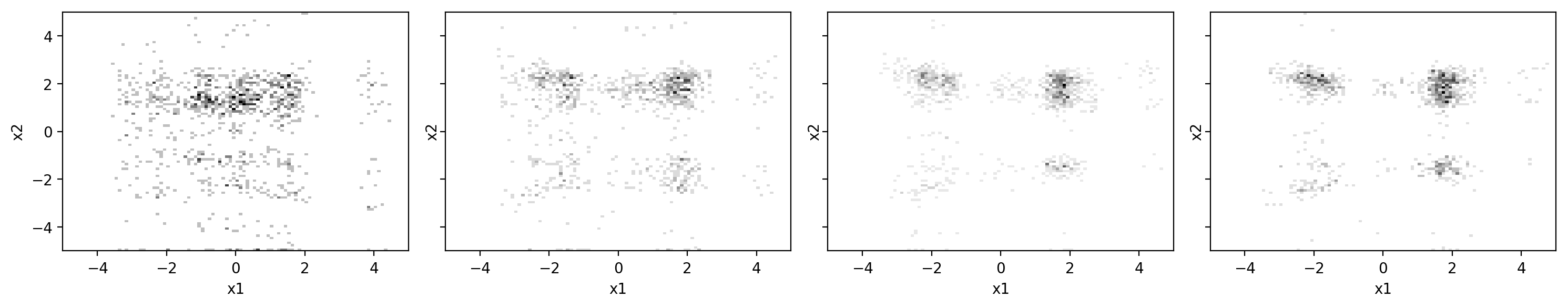}
    \caption{} \par
\end{subfigure}
\caption{Final Distribution of "Synt-2D" function (a) CEM (b) CEM++KDE (c) GACEM on-policy (d) GACEM off-policy}
\label{fig: synt-dist}
\end{figure}

\begin{figure}
\begin{subfigure}{0.22\linewidth}
    \includegraphics[width=\linewidth]{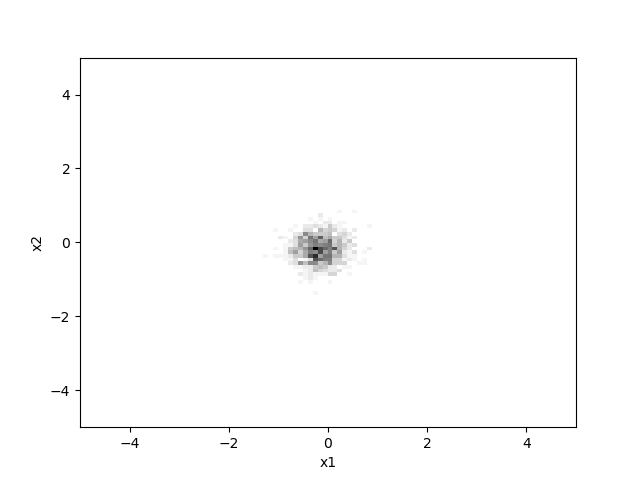}
    \caption{}
\end{subfigure}
\begin{subfigure}{0.22\linewidth}
    \includegraphics[width=\linewidth]{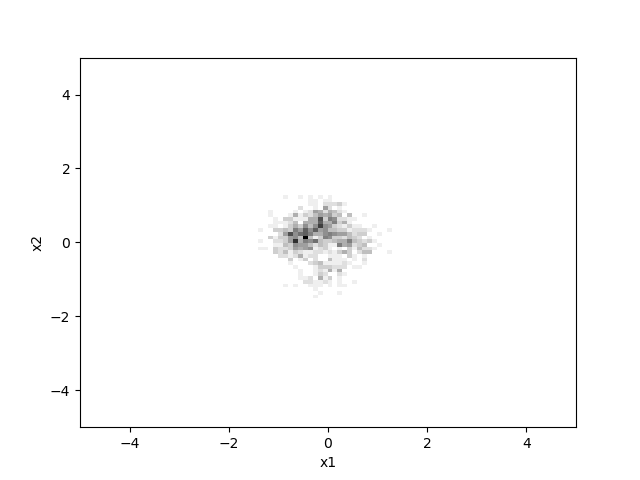}
    \caption{}
\end{subfigure}
\begin{subfigure}{0.22\linewidth}
    \includegraphics[width=\linewidth]{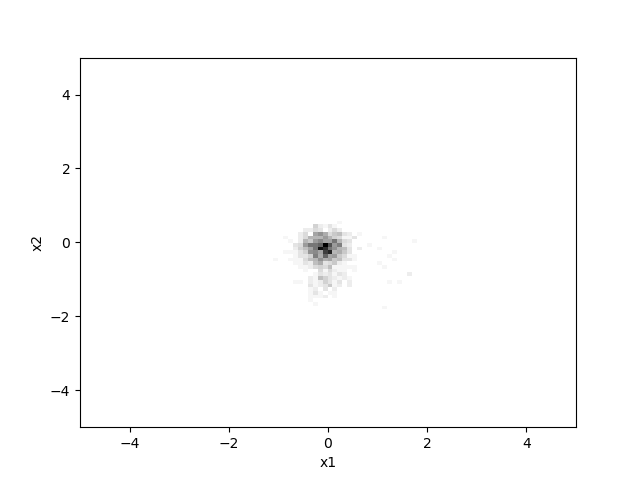}
    \caption{}
\end{subfigure}
\begin{subfigure}{0.22\linewidth}
    \includegraphics[width=\linewidth]{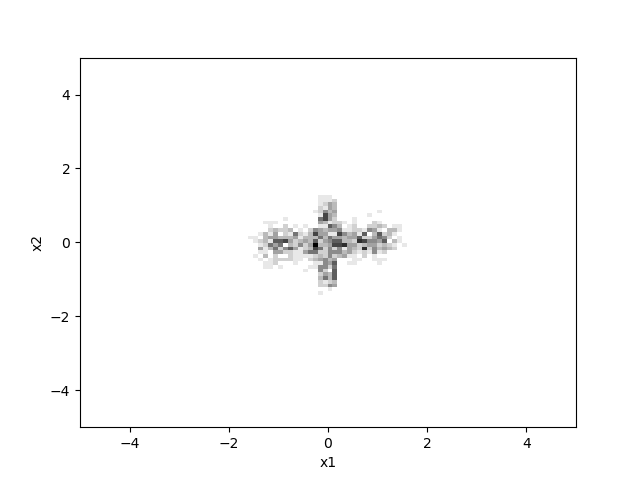}
    \caption{}
\end{subfigure}
\caption{Final Distribution of "Ackley" function (a) CEM (b) CEM++KDE (c) GACEM on-policy (d) GACEM off-policy}
\label{fig: ackley-dist}
\end{figure}

\subsection{Mode discovery}
\label{sec: mode-disc}
 In this section only the "Synt" function is considered. Four algorithms are considered for benchmarks: CEM with fixed variance, CEM++KDE, on-policy GACEM with fixed variance\footnote{Hyper-parameters are in supplementary material}, off-policy GACEM with fixed variance, without the important sampling ratio. As a note, for the off-policy case, not incorporating the importance sampling ratio helps the algorithm converge faster. All of the algorithms are trained on the same amount of data and each iteration corresponds to the same amount of samples evaluated. 

Figure \ref{fig: synt-dist} shows an exemplary development of distribution over time. It is noted that CEM with fixed variance does not have the capability to express all of the modes and therefore only converges to the strongest one, based on the samples at the time. CEM++KDE is able to express the distribution but the choice of bandwidth is arbitrary and the accuracy is not sufficient. The on-policy GACEM can find some number of modes, but is prone to forgetting some modes if sufficient samples are not drawn at each iteration. The off-policy GACEM on the other hand can keep track of all the modes with high accuracy. 

Figure \ref{fig: ackley-dist} shows another example where GACEM out-performs the other algorithms. The "Ackley-2D" function with the given constraint has a cross-like shape for it solution space. The only algorithm that fully expresses the true solution space is the off-policy algorithm proposed by this work.

\begin{figure*}
    \centering
    \begin{subfigure}{0.29\linewidth}
        \includegraphics[width=\linewidth]{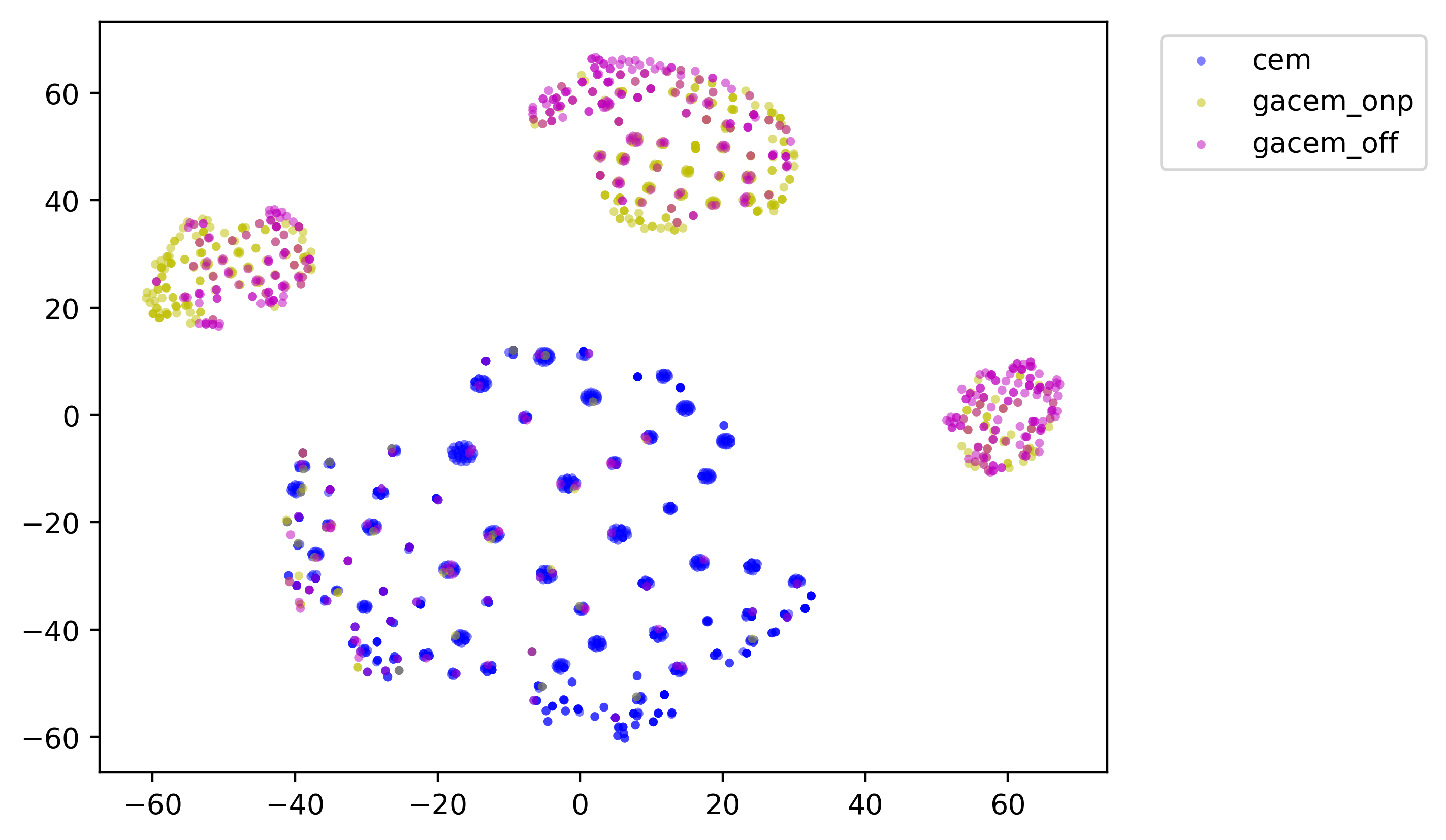}
        \caption{Synt-2D: 4 true clusters}
    \end{subfigure}
    \begin{subfigure}{0.29\linewidth}
    \includegraphics[width=\linewidth]{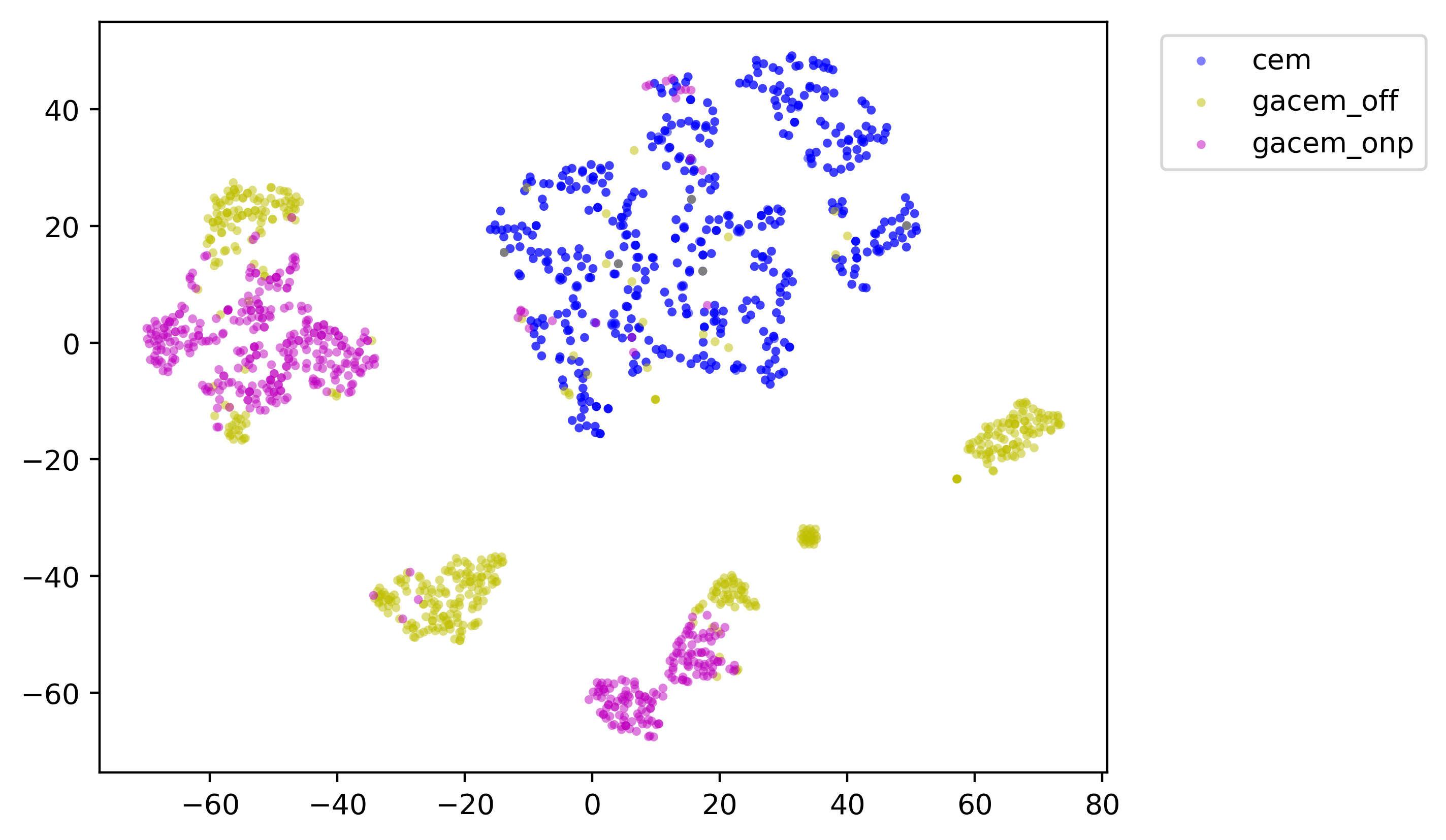}
    \caption{Synt-3D: 8 true clusters}
    \end{subfigure}
    \begin{subfigure}{0.29\linewidth}
    \includegraphics[width=\linewidth]{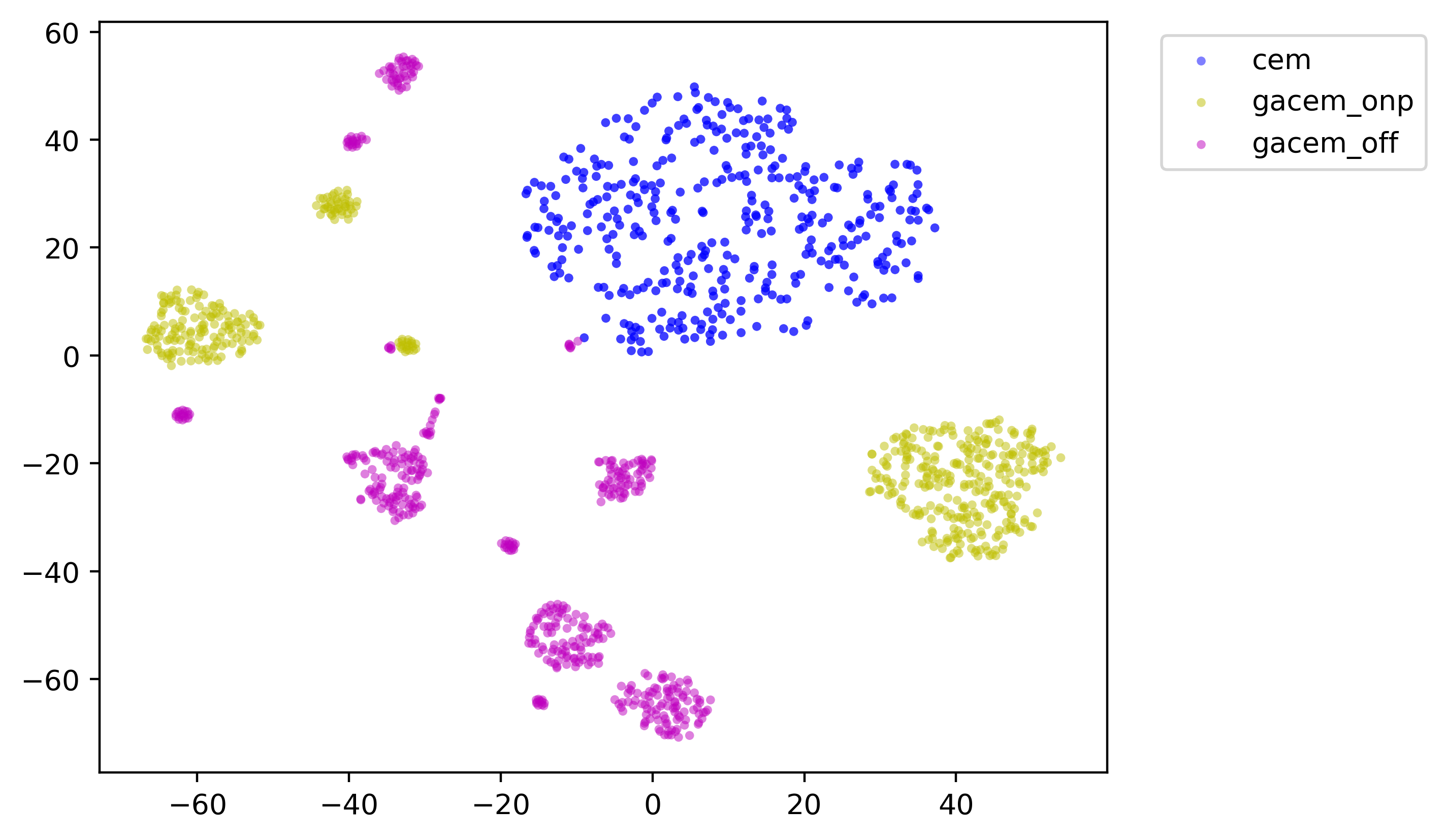}
    \caption{Synt-5D: 32 true clusters}
    \end{subfigure}
    \caption{Examples of t-SNE plots for the solution clouds found by CEM, on-policy GACEM, and off-policy GACEM in optimization of "synt" function}
    \label{fig:tsne}
\end{figure*}

\subsection{Exploration}
This section will investigate the clusters that are formed by the sampled solutions. To do this, we sample the learned solution space for each problem and then look at its t-SNE \cite{van_der_maaten_visualizing_nodate} projection to visualize the clusters. Figure \ref{fig:tsne} shows the results of this experiment for Synt 2D, 3D, and 5D, with 4, 8, and 32 distinctive clusters, respectively, for exemplary seeds. 
From the t-SNE plots it is clear that more than one mode is almost always discovered with GACEM-based algorithms, while with CEM, only one is explored and exploited. It can also be seen that in high dimensions, the GACEM-based algorithms are unable to discover all the modes and only some of them are explored. 

\begin{figure}[h]
    \centering
    \includegraphics[width=\linewidth]{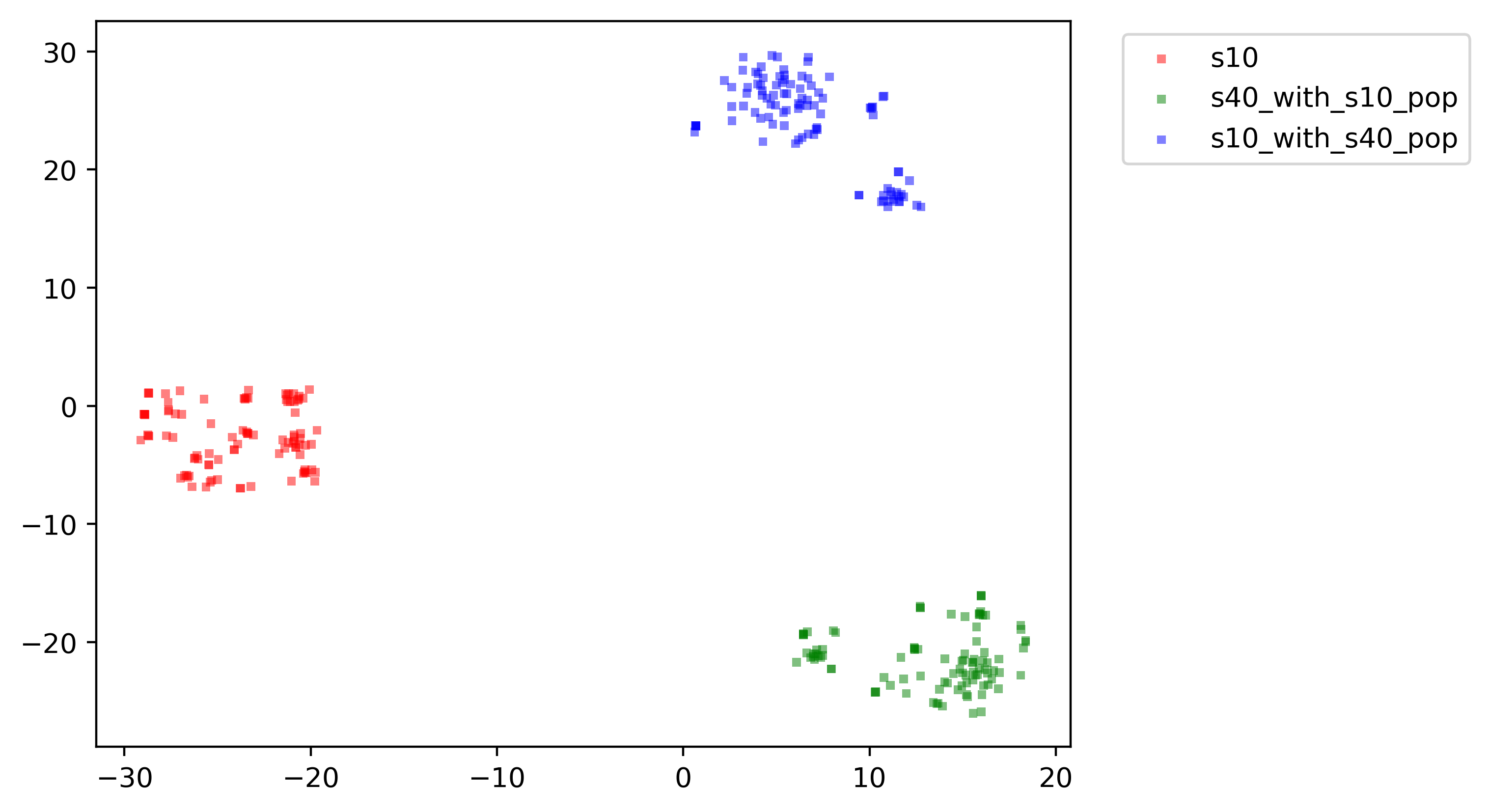}
    \caption{t-SNE plots for solution clouds for three different scenarios: 1. a reference solution based on a random seed (s10), 2. a solution cloud with different initialization but similar population as (1), 3. a solution cloud with similar initialization as (1) but different initial population.}
    \label{fig: init_pop}
\end{figure}

Another question worth asking is how the choice of initial population can affect the results? Is the stochasticity in the initial population the only cause for these differences or the randomness in network's weights and updates also contribute to these different behaviors? In Figure \ref{fig: init_pop}, for Styblinski-20D, a reference solution space is drawn as s10 (red). The other two clusters are differently initialized networks with identical initial populations and similarly initialized networks with different initial populations, respectively. This plot clearly shows that the results of each case is separated in 20D space and therefore, randomness in both initialization of the population and initialization of networks contribute to the differences in mode discoveries. This algorithm does not employ any explicit exploration mechanism different from the normal CEM, and it can indeed get stuck in local optima. However, because it can preserve multi-modal beliefs over solutions space it can out-perform CEM in exploration.

\begin{figure*}
\begin{subfigure}{0.33\linewidth}
\includegraphics[width=\linewidth]{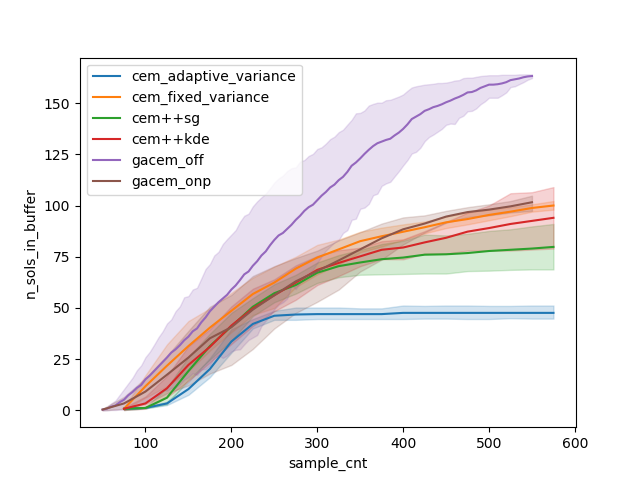}
\par
\includegraphics[width=\linewidth]{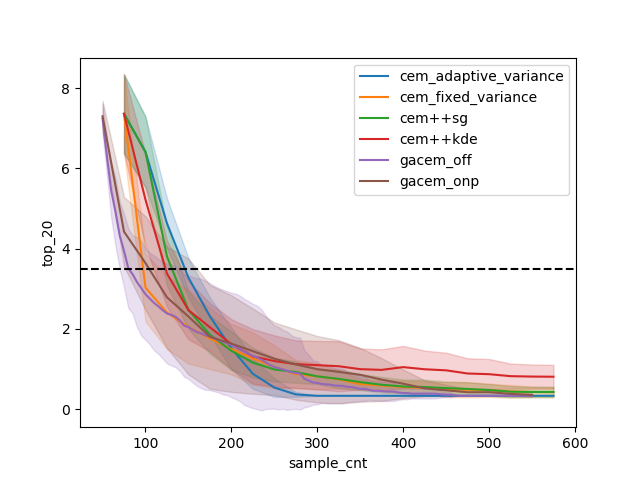}
\centering
\resizebox{\linewidth}{!}{
\begin{tabular}{ccccccc}
    & \textbf{Blue} & \textbf{Orange} & \textbf{Green} &  \textbf{Red} & \textbf{Pink} & \textbf{Brown}\\
    Acc. & 100 & 74 & 83.9 & 63.5 & 71.64 & \textbf{76.7} \\ 
    Ent. & 0 & 2.58 & 1.85 & - & 2.83 & \textbf{2.5}
\end{tabular}
}
\caption{Ackley-2D}
\end{subfigure}
\begin{subfigure}{0.33\linewidth}
\includegraphics[width=\linewidth]{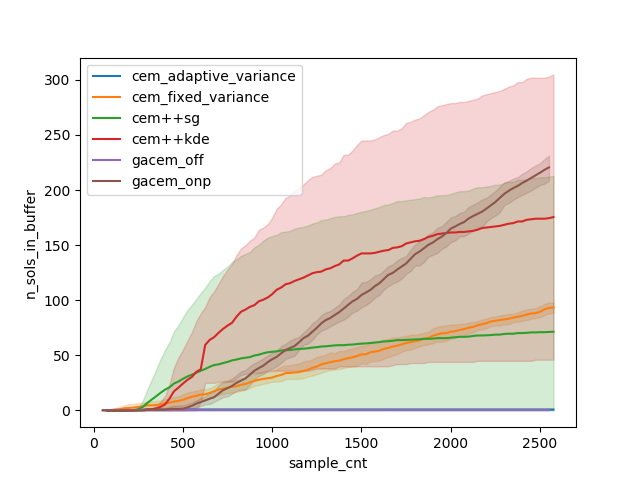}
\par
\includegraphics[width=\linewidth]{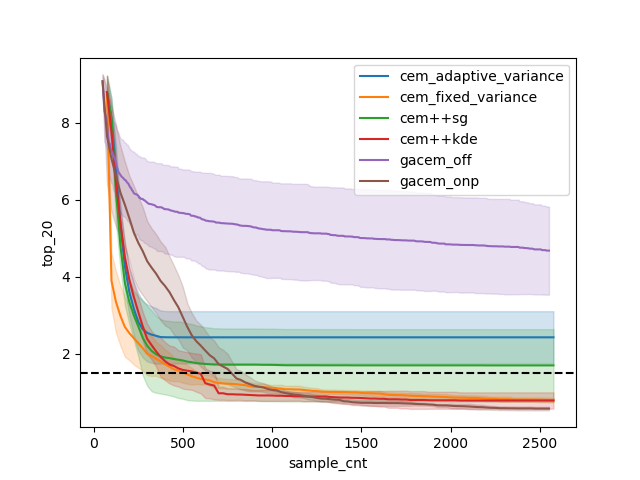}
\centering
\resizebox{0.86\linewidth}{!}{
\begin{tabular}{cccccc}
    \textbf{Blue} & \textbf{Orange} & \textbf{Green} &  \textbf{Red} & \textbf{Pink} & \textbf{Brown}\\
    0 & 3.56 & 8.06 & 9.08 & 0& \textbf{11.86} \\ 
    1.22 & 2.58 & 0.51 & 3.29 & 3.03& \textbf{2.35}
\end{tabular}
}
\caption{Ackley-5D}
\end{subfigure}
\begin{subfigure}{0.33\linewidth}
\includegraphics[width=\linewidth]{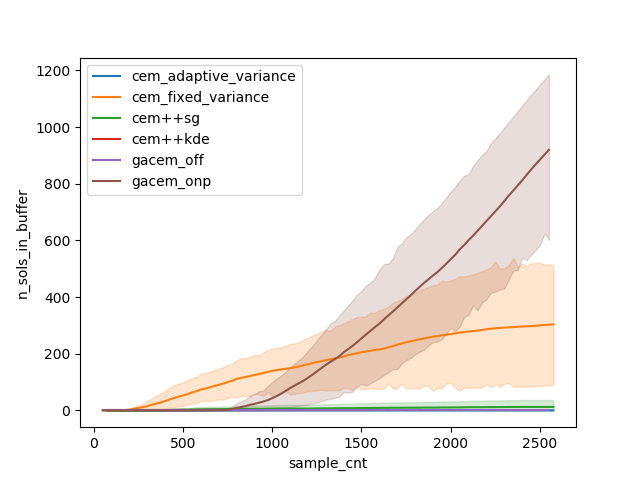}
\par
\includegraphics[width=\linewidth]{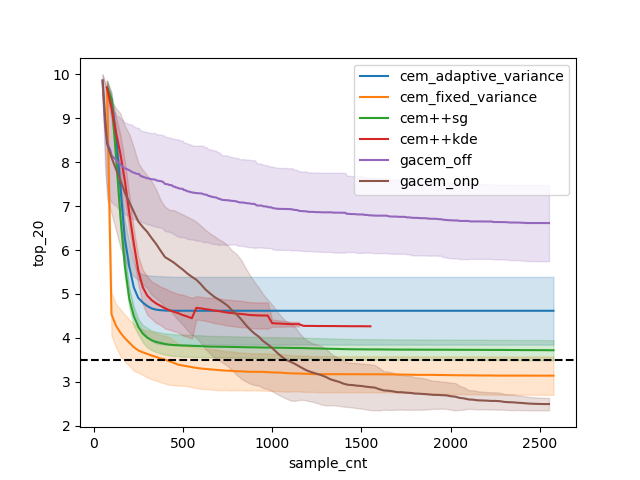}
\centering
\resizebox{0.86\linewidth}{!}{
\begin{tabular}{cccccc}
    \textbf{Blue} & \textbf{Orange} & \textbf{Green} &  \textbf{Red} & \textbf{Pink} & \textbf{Brown}\\
    0 & 4.3 & 0.3 & 0 & 0 & \textbf{73} \\ 
    0.86 & 2.59 & 1 & 4.4 & 3.37 & \textbf{2.48}
\end{tabular}
}
\caption{Ackley-20D}
\end{subfigure}
\caption{Performance of All algorithms on Ackley function with various dimensions.}
\label{fig: ackley-perf}
\end{figure*}

\subsection{Scale with dimensions}
\label{sec: dim}
This section will investigate how each algorithm's performance is affected by an increase in the problem's dimensionality. For each problem with a given difficulty, all algorithms are run and the number of satisfying solutions are plotted as a function of total number of samples evaluated. Each curve in the figure in the top row of figure \ref{fig: ackley-perf} shows the average and confidence interval of this metric across different initial seeds.
The average value of the top-20 solutions is also plotted in the second row. For quantitative measurements we consider two metrics: accuracy, which is defined as the percentage of samples in a batch, obtained from the distribution after training, that satisfy the constraint, and entropy per dimension which is the estimated entropy of the learned solution space distribution divided by number of dimensions. An ideal distribution would have an accuracy of 100 percent and an entropy of $log(V)$ where $V$ is the volume of solution space ($log(N)$ in the discrete case where N is the number of available solutions). The entropy's magnitude is $\mathcal{O}(d)$ (where $d$ is number of dimensions) and therefore, dividing it by $d$ will allow comparison across different dimensional versions of the same problem. 

The following algorithms are considered as benchmarks: CEM with adaptive variance, CEM with fixed variance (where variance is chosen using cross-validation), CEM++SG, CEM++KDE, GACEM with on-policy sampling and GACEM with off-policy updates without the importance sampling ratio. 

The results show that while the off-policy GACEM works significantly better in finding the shape of the true distribution in low dimensions (section \ref{sec: mode-disc}), it cannot be scaled to higher dimensions. The on-policy GACEM instead, consistently has a better solution diversity than the other algorithms (because it can explore multiple-modes at the same time). In terms of the top-20 average, CEM-based algorithms have more momentum in exploitation, while their performance can get saturated in a sub-optimal region because of less exploration. The results of a similar experiment with other functions are detailed in the supplementary material.

The results show that among CEM-based approaches CEM with fixed variance is superior in terms of both solution diversity and top-20 performance. The on-policy GACEM out-performs CEM with fixed variance in terms of diversity of solutions in all settings. The performance of top-20 designs in on-policy GACEM reaches to levels beyond CEM with fixed variance but requires more evaluations. In 2D case the off-policy GACEM out-performs all other algorithms, however, empirically, it lacks scalablity to higher dimensions.
\section{Issues and Future Direction}
The followings are future directions that the current algorithm can be improved upon:
\begin{itemize}
    \item The off-policy version with importance sampling does not work in practice in higher dimensions, but it has shown better capabilities in finding the modes in lower dimensions that other algorithms. More investigation is required to understand the issue and to provide answers. 
    \item The exploration is limited. One possible work for the future is to see if existing exploration encouragements in Reinforcement Learning (i.e. adding an encouragement term for exploring novel samples to $w(x)$) can be adapted to this type of optimization algorithm.
    \item The algorithm shows less momentum towards the solution space compared to CEM, mainly because of the usage of stochastic gradient descent in optimizing the parameters of distribution. The authors believe that better architectures and more stable gradient estimations can bridge the gap between CEM and GACEM, in terms of sample efficiency.
\end{itemize}
\section{Conclusion}
In this work, a generalized cross-entropy method for solving rare event constraint satisfaction problems has been proposed. The generalization emerges from the utilization of auto-regressive probabilistic models to model the distribution over solution space. This model can be trained with the on-policy REINFORCE algorithm and scales well to optimizing high dimensional problems. Compared to CEM it is capable of solving multi-modal optimization problems, and shows better diversity in the solution space.

\bibliography{main}
\bibliographystyle{icml2020}

\end{document}


\begin{figure*}[!h]
\begin{center}
   \large\textbf{Supplementary Material}
\end{center}

\section{Performance of GACEM on other test functions}
\begin{subfigure}{0.32\textwidth}
\includegraphics[width=\linewidth]{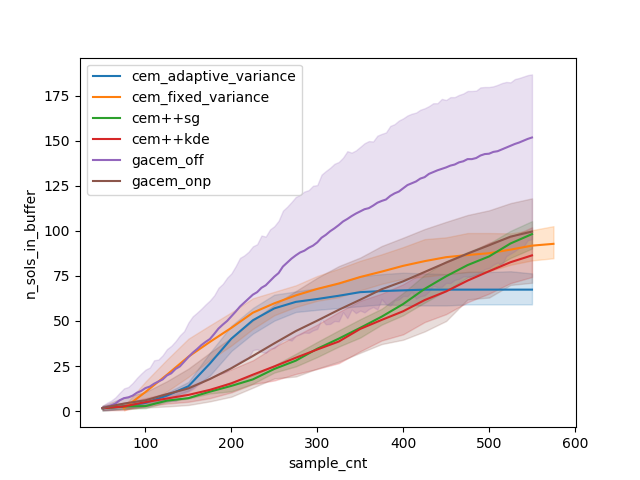}
\par
\includegraphics[width=\linewidth]{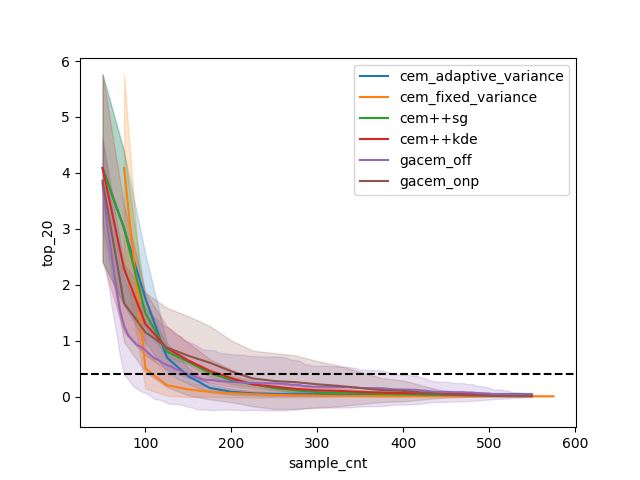}
\centering
\resizebox{\linewidth}{!}{
\begin{tabular}{ccccccc}
    & \textbf{Blue} & \textbf{Orange} & \textbf{Green} &  \textbf{Red} & \textbf{Pink} & \textbf{Brown}\\
    Acc. & 100 & 67.3 & 65.62 & 53.36 & 73.8 & \textbf{73.06} \\ 
    Ent. & 0.14 & 2.58 & 2.91 & 3.1 & 2.86 & \textbf{2.61}
\end{tabular}
}
\caption{Levy-2D}
\end{subfigure}
\begin{subfigure}{0.32\textwidth}
\includegraphics[width=\linewidth]{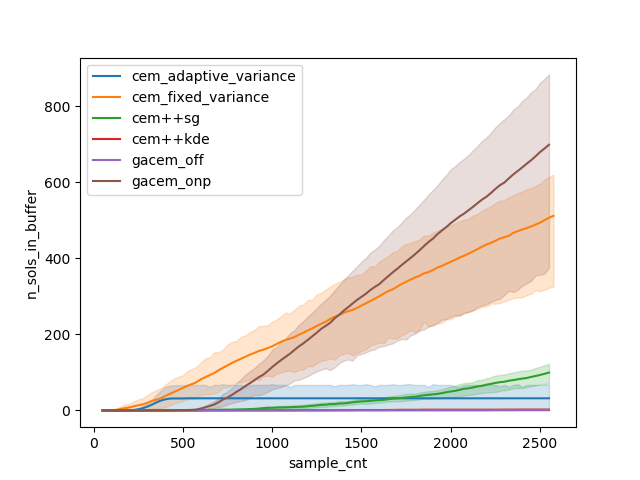}
\par
\includegraphics[width=\linewidth]{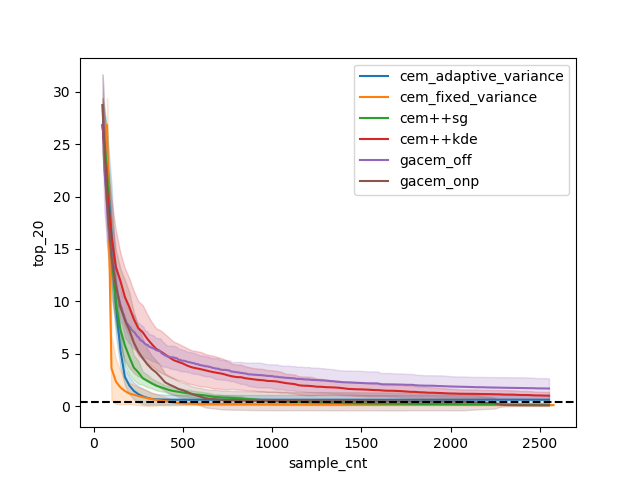}
\centering
\resizebox{0.86\linewidth}{!}{
\begin{tabular}{cccccc}
    \textbf{Blue} & \textbf{Orange} & \textbf{Green} &  \textbf{Red} & \textbf{Pink} & \textbf{Brown}\\
    0 & 21.88 & 10.9 & 0.14 & 0.08 & \textbf{38.38} \\ 
    1.90 & 2.58 & 2.98 & 4.00 & 2.77 & \textbf{2.37}
\end{tabular}
}
\caption{Levy-5D}
\end{subfigure}
\begin{subfigure}{0.32\textwidth}
\includegraphics[width=\linewidth]{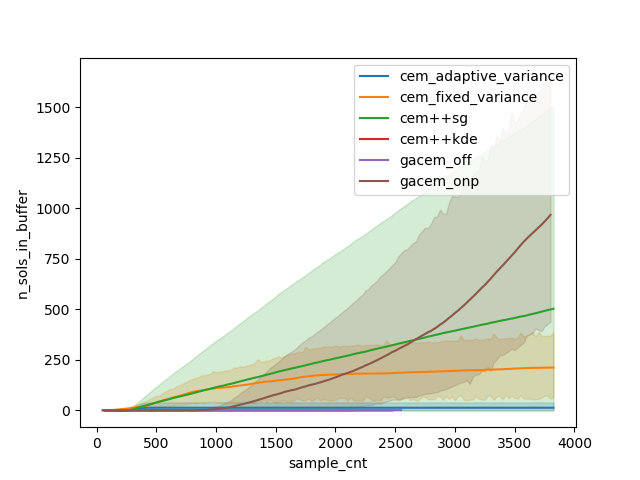}
\par
\includegraphics[width=\linewidth]{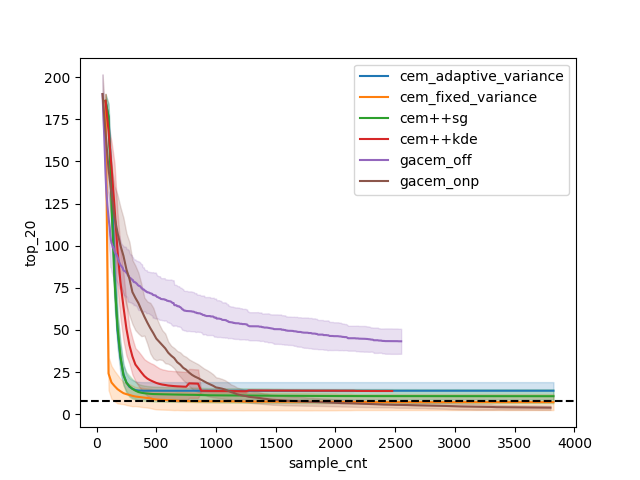}
\centering
\resizebox{0.86\linewidth}{!}{
\begin{tabular}{cccccc}
    \textbf{Blue} & \textbf{Orange} & \textbf{Green} &  \textbf{Red} & \textbf{Pink} & \textbf{Brown}\\
    0 & 4.3 & 0.3 & 0 & 0 & \textbf{73} \\ 
    0.86 & 2.59 & 1 & 4.4 & 3.37 & \textbf{2.48}
\end{tabular}
}
\caption{Levy-20D}
\end{subfigure}
\caption{Performance of All algorithms on Levy function with various dimensions.}
\label{fig: ackley-perf}
\begin{subfigure}{0.32\textwidth}
\includegraphics[width=\linewidth]{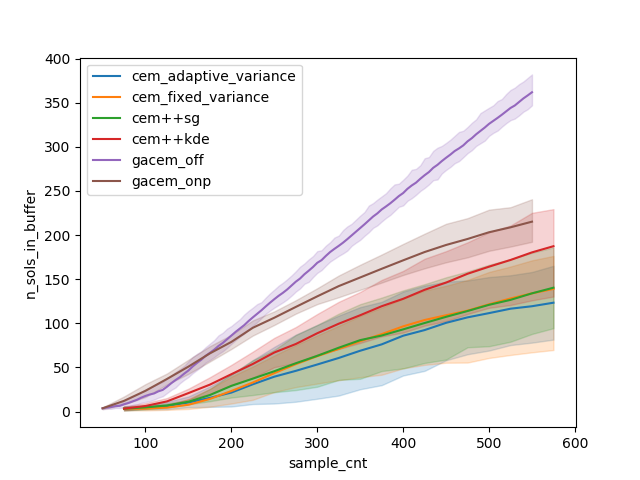}
\par
\includegraphics[width=\linewidth]{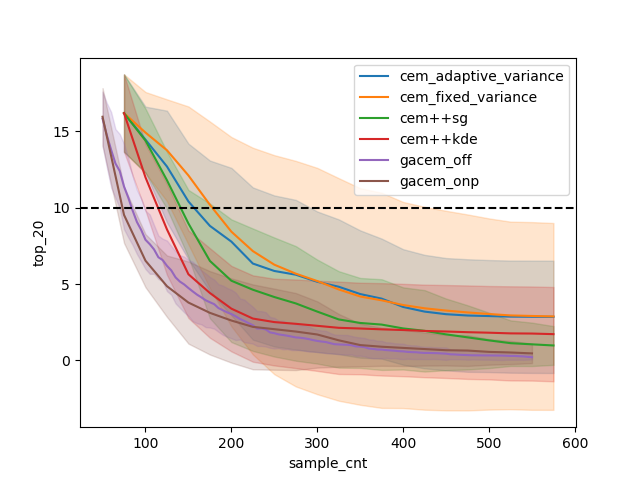}
\centering
\resizebox{\linewidth}{!}{
\begin{tabular}{ccccccc}
    & \textbf{Blue} & \textbf{Orange} & \textbf{Green} &  \textbf{Red} & \textbf{Pink} & \textbf{Brown}\\
    Acc. & 0 & 0.66 & 13.08 & 0 & 0 & \textbf{65.54} \\ 
    Ent. & 0 & 2.59 & 1.47 & 4.6 & 3.49 & \textbf{2.43}
\end{tabular}
}
\caption{Styblinski-2D}
\end{subfigure}
\begin{subfigure}{0.32\textwidth}
\includegraphics[width=\linewidth]{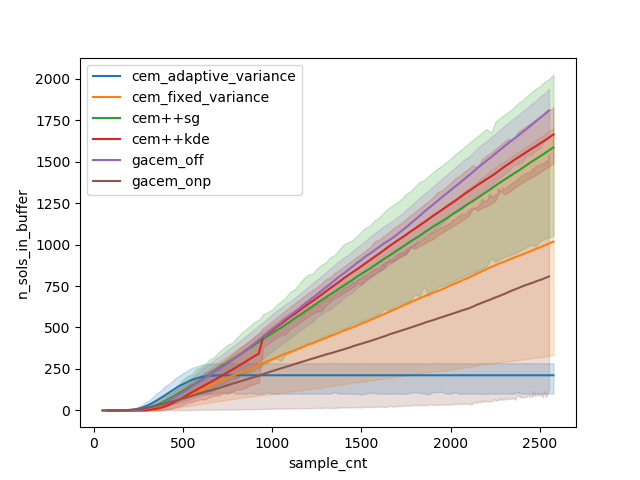}
\par
\includegraphics[width=\linewidth]{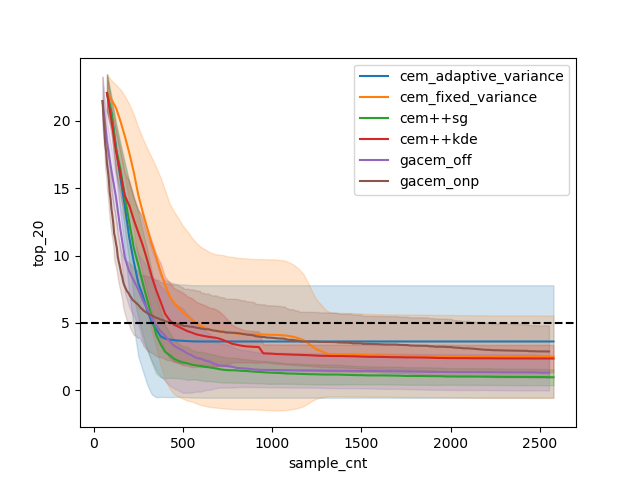}
\centering
\resizebox{0.86\linewidth}{!}{
\begin{tabular}{cccccc}
    \textbf{Blue} & \textbf{Orange} & \textbf{Green} &  \textbf{Red} & \textbf{Pink} & \textbf{Brown}\\
    0 & 42.26 & 77.72 & 68.84 & 90.16 & \textbf{43.62} \\ 
    1.08 & 2.59 & 2.23 & 2.50 & 3.37 & \textbf{2.52}
\end{tabular}
}
\caption{Styblinski-5D}
\end{subfigure}
\begin{subfigure}{0.32\textwidth}
\includegraphics[width=\linewidth]{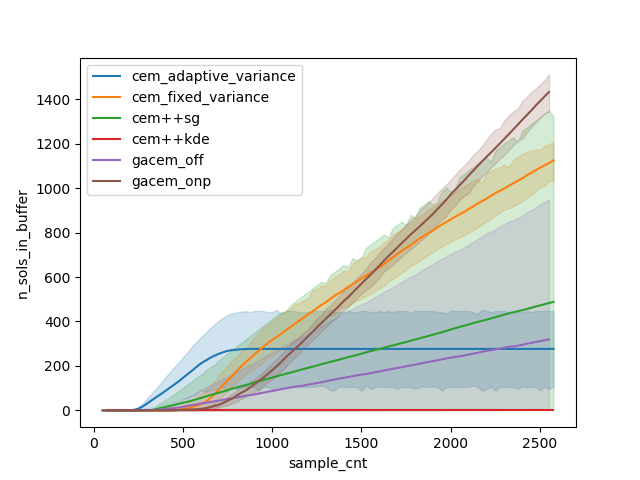}
\par
\includegraphics[width=\linewidth]{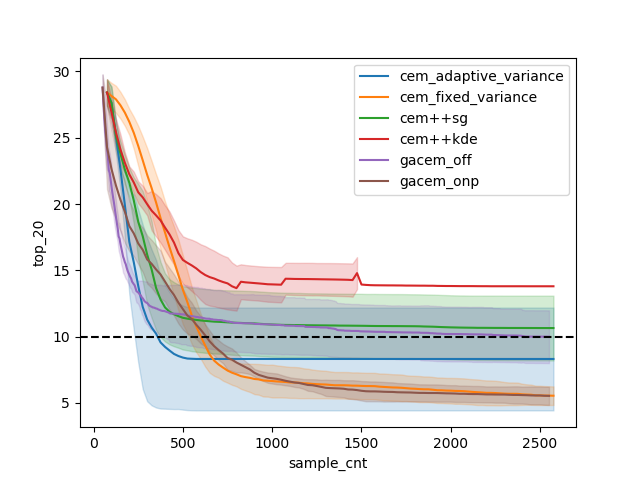}
\centering
\resizebox{0.86\linewidth}{!}{
\begin{tabular}{cccccc}
    \textbf{Blue} & \textbf{Orange} & \textbf{Green} &  \textbf{Red} & \textbf{Pink} & \textbf{Brown}\\
    0 & 46.16 & 35.05 & 0 & 17.08 & \textbf{77.84} \\ 
    0 & 2.59 & 0.66 & 4.62 & 3.18 & \textbf{2.54}
\end{tabular}
}
\caption{Styblinski-20D}
\end{subfigure}
\caption{Performance of all algorithms on Styblinski function with various dimensions.}
\label{fig: ackley-perf}

\end{figure*}
\newpage

\begin{figure}[!thp]
\section{Hyper-parameters}
\centering
    \begin{tabular}{l|c|l}
         \textbf{Name} & \textbf{Value} & \textbf{Explanation}  \\
         \hline
         nr\_mix & 40 & Number of Gaussian Mixtures\\
         $\sigma_{GACEM}$ & 0.05 &  Variance used in GACEM with fixed variance \\
         $\mathbf{\beta}$ & 10 & Control parameter for entropy 
        \\
         hidden layers & 3$\times$fc[100] & The NN has 3 hidden fully connected layers with 100 units on each \\
         learning rate & 5e-3 &  Learning rate for optimization \\
         $n_{init}$ & 50 & Number of initial samples obtained randomly\\
         $n_s$ & 25 & Number of samples evaluated at each iteration \\
         $n_e$ & 10 & Number of epochs that NN is trained for in each iteration \\
         $b_s$ & 16 & Batch size used when training NN in each iteration \\
         \hline
         CEM elite percentile & \%40 & The percentile which CEM uses to determine the elite individuals \\
         $\sigma_{CEM}$ & 0.05 & Variance used in CEM with fixed variance \\
    \end{tabular}
    \caption{List of hyper-parameters used for all algorithms for all test functions.}
    \label{tab:my_label}
\end{figure}